%% file: main.tex
\documentclass[conference]{IEEEtran}
\IEEEoverridecommandlockouts
\usepackage{cite}
\usepackage{amsmath,amssymb,amsfonts}
\usepackage{algorithmic}
\usepackage{booktabs}
\usepackage{graphicx}
\usepackage{textcomp}
\usepackage{xcolor}
\usepackage{soul}
\usepackage{multirow}
\usepackage{array}
\usepackage{caption}
\def\BibTeX{{\rm B\kern-.05em{\sc i\kern-.025em b}\kern-.08em
    T\kern-.1667em\lower.7ex\hbox{E}\kern-.125emX}}
\usepackage{comment}

\newcommand{\kevin}[1]{{\textbf{\textcolor{orange}{[Kevin: #1]}}}}

\setstcolor{red}

\newcommand\copyrighttext{%
  \footnotesize \textcopyright \the\year{} IEEE. Personal use of this material is permitted. Permission from IEEE must be obtained for all other uses, including reprinting/republishing this material for advertising or promotional purposes, collecting new collected works for resale or redistribution to servers or lists, or reuse of any copyrighted component of this work in other works.}

\begin{document}

\title{GenAI for Social Work Field Education: Client Simulation with Real-Time Feedback 
}

\author{\IEEEauthorblockN{James Sungarda}
\IEEEauthorblockA{\textit{School of Computing and Data Science} \\
\textit{The University of Hong Kong}\\
Hong Kong, China \\
u3632598@connect.hku.hk}
\and
\IEEEauthorblockN{Hongkai Liu}
\IEEEauthorblockA{\textit{Dept. of Mathematics} \\
\textit{The University of Hong Kong}\\
Hong Kong, China \\
u3612776@connect.hku.hk}
\and
\IEEEauthorblockN{Zilong Zhou}
\IEEEauthorblockA{\textit{School of Computing and Data Science} \\
\textit{The University of Hong Kong}\\
Hong Kong, China \\
zzl0712@connect.hku.hk}
\and
\IEEEauthorblockN{Tien-Hsuan Wu}
\IEEEauthorblockA{\textit{Faculty of Engineering} \\
\textit{The University of Hong Kong}\\
Hong Kong, China \\
wukevin@hku.hk}
\and
\IEEEauthorblockN{Johnson Chun-Sing Cheung}
\IEEEauthorblockA{\textit{Dept. of Social Work and Social Administration} \\
\textit{The University of Hong Kong}\\
Hong Kong, China \\
cjcs@hku.hk}
\and
\IEEEauthorblockN{Ben Kao}
\IEEEauthorblockA{\textit{School of Computing and Data Science} \\
\textit{The University of Hong Kong}\\
Hong Kong, China \\
kao@cs.hku.hk}
}

\maketitle

\copyrighttext

\begin{abstract}
Field education is the signature pedagogy of social work, yet providing timely and objective feedback during training is constrained by the availability of instructors and counseling clients.
In this paper, we present SWITCH, the Social Work Interactive Training Chatbot.
SWITCH integrates realistic client simulation, real-time counseling skill classification, and a Motivational Interviewing (MI) progression system into the training workflow.
To model a client, SWITCH uses a cognitively grounded profile comprising static fields (e.g., background, beliefs) and dynamic fields (e.g., emotions, automatic thoughts, openness), allowing the agent’s behavior to evolve throughout a session realistically.
The skill classification module identifies the counseling skills from the user utterances, and feeds the result to the MI controller that regulates the MI stage transitions.
To enhance classification accuracy, we study in-context learning with retrieval over annotated transcripts, and a fine-tuned BERT multi-label classifier.
In the experiments, we demonstrated that both BERT-based approach and in-context learning outperforms the baseline with big margin.
SWITCH thereby offers a scalable, low-cost, and consistent training workflow that complements field education, and allows supervisors to focus on higher-level mentorship.

\end{abstract}


\input{1-introduction}
\input{2-related}

\input{3-system}
\input{4-classification}
\input{5-experiment}

\input{6-deploy}
\input{7-conclusion}

\section*{Acknowledgment}
This project is supported by the HKU Teaching Development Grant (project title: A Cross-disciplinary Application of Generative Artificial Intelligence (GenAI) in Innovating Simulation-based Learning of Social Casework), the Tam Wing Fan Innovation Fund, and the Philomathia Foundation Innovation Fund.
This project is approved by the HKU Human Research Ethics Committee (EA250449).

\bibliographystyle{ieeetran}
\bibliography{references}

\end{document}

%% file: 1-introduction.tex
\section{Introduction}

The Council on Social Work Education (CSWE) designated \emph{field education} as the ``signature pedagogy'' of social work education\cite{cswe2015}.
In this model, trainees develop essential skills by engaging in conversations with real clients. 
They work under a social agency and providing support and therapy for clients of various backgrounds\cite{fwswe,fwswe_india}. 
Field work provides students with opportunities to integrate theory and practice skills, and to experience the expression of professional values in real-world social contexts \cite{cswe2015,fwswe_india}.

A key component in the execution of field training is the assessment and feedback provided by field instructor.
The instructor's role is to guide these sessions and give feedback by highlighting the social work skills the trainees 
have not used, blind spots, or skills that were not used properly 
during interactions with the client\cite{fwswe,bogo2010}.
To structure these interactions and the feedback process, both students and supervisors utilize various established social intervention frameworks. 
This serves as the grounded theoretical basis in which the social workers' apply their skills from\cite{hohman2024mifield}. 


Recent advancements in Large Language Models (LLMs) enable the development of realistic chatbots\cite{tamoyan2025roleplay},
which can be used to create simulated training platforms that can streamline in-classroom social work training and complement field education. 
An AI-driven comprehensive training suite offers numerous inherent advantages:


\noindent$\bullet$ While the personalized guidance of a field instructor is invaluable, traditional performance assessment often relies on subjective evaluation\cite{bogo2010}. 
An automated solution can streamline the process by introducing an system that captures key performance metrics objectively and instantly. 
This technology does not replace the instructor but rather empowers them. 
Instead it provides a foundation of objective, data-driven insights, and allows instructors to focus on delivering high-level, personalized feedback more efficiently.
    
 \noindent$\bullet$ AI-simulated clients can provide students with consistent exposure to a diverse range of client scenarios\cite{fwswe, bogo2010}.
    Crucially, this includes scenarios where it might jeopardize a real client’s well being.
    Having a safe platform that trainees can use to handle vulnerable cases mitigates the risks of potentially unwanted effects of handling these vulnerable clients.
    
 \noindent$\bullet$ The availability of a low-cost, on-demand solution increases the efficiency and accessibility of social work education itself. 
 Organizing training sessions are expensive and require the coordination of students and instructors\cite{tanga2012lesotho, judy2018rural}.
 A digital system is low cost and hence offer an on-demand solution that can be used anytime and deployed in most circumstances where field training might be unfeasible.


Previous works have created general conversational client chatbots in the social work context\cite{yang2025chatbot, wang2024patient, chan2023chatbot}.
The key ideas introduced in these papers, such as the modelling of an internal cognitive model of the client\cite{wang2024patient} and the appropriate structure for a \textit{persona} for LLMs\cite{yang2025chatbot}, have enabled for realistic LLM-based simulation of social work clients. 


Existing LLM-based systems in social work primarily simulate client conversations. 
However, they stop short of an end-to-end training workflow that links scenarios to competencies, enables transparent and consistent assessment, and supports instructor feedback and longitudinal development. 
This gap is consequential given supervision constraints. 
Bogo\cite{bogo2010} noted that limited supervisor time makes manual review of recordings impractical and recommended integrating observation and feedback within the training process. 
A more comprehensive, structured training solution is an implementation of this idea, lowering the costs of organizing practice and improving the efficiency and consistency of assessment.


In this paper, we introduce \underline{\bf S}ocial \underline{\bf W}ork \underline{\bf I}nteractive \underline{\bf T}raining \underline{\bf Ch}atbot (SWITCH). 
The system's functionality is threefold. 
First, create a structured social work training workflow by using Motivational Interviewing (MI) framework.
MI is a stage-based framework, with each stage representing the willingness of the client to comply with the social worker and to change their behavior. 
The system evaluates the user's skills based on social work skills demonstrated by the trainee, which are identified by a separate skills classification model.
It controls the progression of the training session by deciding when to advance the client through the MI stages.
Second, it incorporates an accurate social work skills classification model.
The skills classified by this model serves as a key metric in which personalized feedback are based on, as well as being the data provider for the client behavior progression system. 
Third, create a diverse set of client simulations modeled on real-world cases, including sensitive cases where a real client might be too vulnerable and risky for the trainee to handle. 
These simulations aim to reflect the diversity and complexity of actual practice. 
\label{sec:intro}


%% file: 2-related.tex
\section{Related Work}
\label{sec:related}

\subsection{AI-Powered Simulated Clients for Training}

Recent advances in conversational AI have enabled realistic simulation of both practitioners and clients for mental health training. While studies have successfully simulated social workers themselves~\cite{xie2024fewshot,he2023conversational}, our work focuses on the complementary challenge of simulating clients. 
ClientBot~\cite{tanana2019clientbot} introduced this approach in 2019, achieving a 91\% increase in counselor reflections during training sessions.
PATIENT-$\psi$~\cite{wang2024patient} introduced cognitive modeling to create psychologically realistic patient simulations. 
In their implementation, they define \emph{cognitive model} to be a set of fields that they claim allow the LLM to accurately simulate the behavior of a persona.
They proved that inserting these fields into the system prompt of the LLM allowed for more accurate client simulations.

Scaffolding Empathy~\cite{steenstra2025scaffolding} employed multi-agent architecture with performance visualizations. 
Recently, Yang et al.~\cite{yang2025chatbot} implemented MI principles in AI agents for consistent client simulation. However, these systems were designed for general mental health contexts and lack comprehensive skill-specific feedback. 
Building on these directions, we integrate CBT-based cognitive modeling tailored for social work, introduce an openness level that evolves across MI stages, and unify these elements into a system designed to support social work training.

\subsection{In-Context Learning}
In-context learning (ICL) has emerged as a transformative paradigm for NLP systems, enabling adaptation to new tasks without gradient updates. Brown et al.\cite{brown2020languagemodelsfewshotlearners} demonstrated with GPT-3 that large language models can perform new tasks from just a few examples, establishing the foundation for prompt-based few-shot learning. Madotto et al.~\cite{madotto2021fewshotbotpromptbasedlearning} demonstrated that prompt-based few-shot learning eliminates gradient-based fine-tuning in dialogue systems, benchmarking LMs across nine response generation tasks using only examples in the LM context as the learning source. For classification with extensive label spaces, Milios et al.~\cite{milios2023incontextlearningtextclassification} addressed context window limitations by employing pre-trained dense retrieval models, providing partial label space views that achieved state-of-the-art performance on intent classification without fine-tuning.

\subsection{Counseling Skills Classification}
Traditional classification methods laid the groundwork for automated counseling skill analysis before the transformer revolution. Early computational approaches relied on statistical and machine learning techniques. Althoff et al.~\cite{althoff2016largescaleanalysiscounselingconversations} applied Hidden Markov Models to analyze counseling conversations, identifying distinct conversation stages and demonstrating that statistical models could predict therapeutic outcomes from early conversation dynamics. 
Automated classification of therapeutic conversations has shown promise using transformer-based approaches. MentalBERT~\cite{ji2021mentalbert} demonstrated domain-specific pretraining advantages, while Working Alliance Transformer~\cite{lin2024working} created custom architectures for analyzing therapy sessions. However, these approaches focused primarily on the classification of client response. 
We first introduce skill classification as a separate module that identifies counseling skills in real-time, enabling instant feedback and training control. Using fine-tuned BERT combined with in-context learning, our modular design transforms skill assessment from post-hoc analysis into an active training component that guides skill development.

%% file: 3-system.tex
\section{System Design and Architecture Work}
\label{sec:system}

Field training in social work consists of three stages: practice, observation and feedback\cite{cswe2015, bogo2010}.
In SWITCH, the goal is to automate the three stages of field training, easing the process of the field education practice loop.

\begin{figure*}[htbp]
    \centering
    \includegraphics[width=0.8\linewidth]{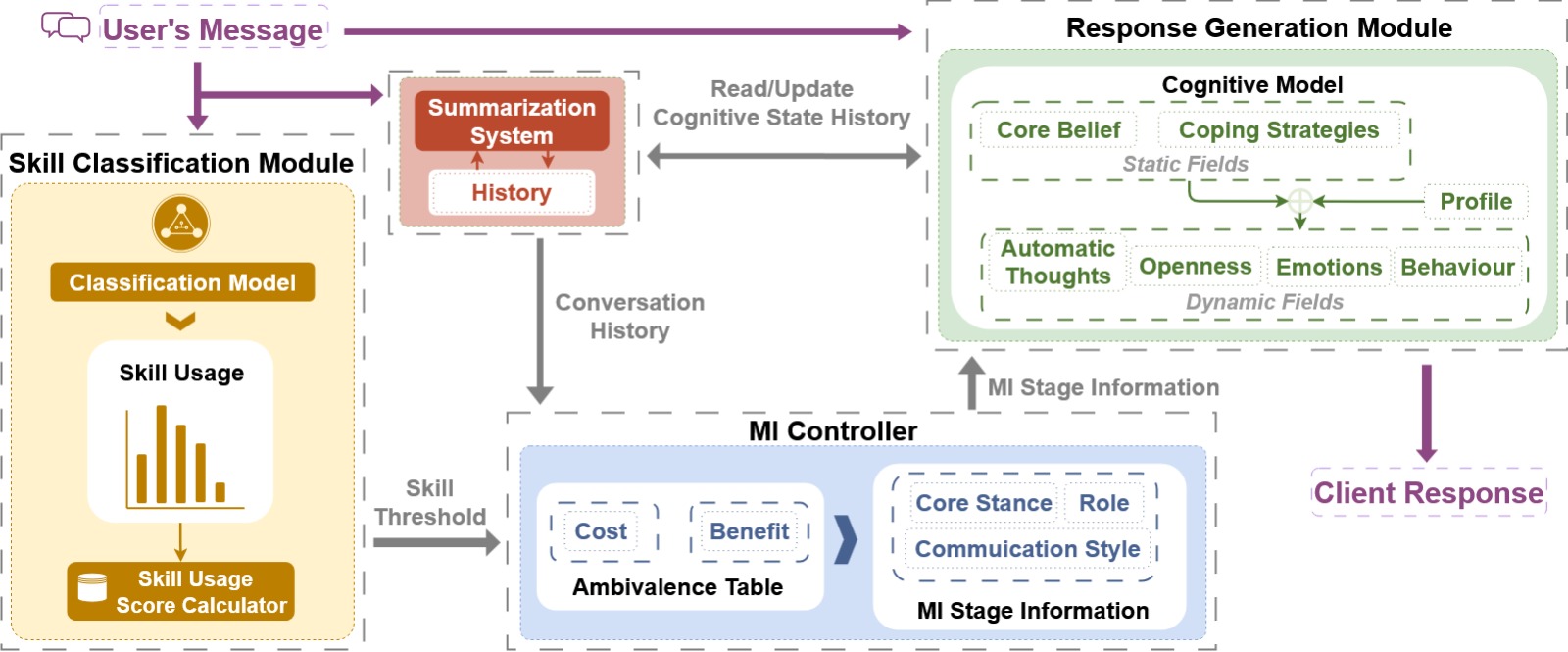}
    \caption{Workflow of SWITCH}
    \label{fig:switch_workflow}
\end{figure*}

Figure~\ref{fig:switch_workflow} shows the workflow of SWITCH.
When a user provides a message, the skill classification module identifies the counseling skills demonstrated by the user.
The response generation module updates the simulated client’s cognitive state and produces a corresponding reply.
Based on the conversation, the Motivational Interviewing (MI) controller determines if the client should progress to the next MI stage.
In this section, we detail the design of SWITCH, and how we effectively automate the processes of field work training.

    
\subsection{Motivational Interviewing (MI)}

Traditional field work with real clients are guided by various \textit{conceptual frameworks}\cite{bogo2010}.
This acts as a ``knowledge base'' which the trainees refer to when applying various social work skills, 
bridging the theory they have received in classroom and the practice of communicating with the clients themselves\cite{bogo2010, fwswe}.
Simulating the methodology of field work with real clients, in this section we discuss we integrated the Motivational Interviewing (MI) framework to the training workflow.



Motivational Interviewing is an effective counseling approach designed to strengthen individual's motivation and commitment to change\cite{richardson2004mi, hohman2024mifield, miller2012motivational}.
MI identifies five stages of change: {\it pre-contemplation}, {\it contemplation}, {\it preparation}, {\it action}, and {\it maintenance}. 
In each {\it MI stage}, clients exhibit distinct thoughts, emotions, openness, and behaviors.
For the purpose of SWITCH, we consider the first three stages only, as the final two stages are typically seen in post-counseling follow-ups and fall beyond the scope of training chatbots.

The key difference in this stage-based approach is the addition of \emph{stage info}. 
Stage info is a stage-specific structured set of instructions and descriptive cues that guides the client simulation model 
in alignment with the characteristic behaviors associated with each MI stage\cite{bogo2010}: what the client is ready to discuss (problem, solutions, or plan) and what they will hold off on.
A general example of MI stage information can be seen on Table I.
As these set of instructions are mapped to each individual stage, it only changes when the MI stage progresses.
The current stage info instructions will be inserted in the prompt for the response generation LLM.
The response generation process will be explained in further detail in Section~\ref{sec:profile}.

\input{figs/mi}



The progression of the stages are controlled by two factors: a \emph{skill score} determined by the number and type of skills used in the conversation, 
and a \emph{MI controller} model that analyzes the context of the current MI stage and verifies if the current MI goals are met.




\noindent\textbf{[Skill Score]}
For each skill $l_j$, we compute its frequency of occurrence ($n_j$). 
At each stage $t$, a weight $w_j^t\in \{0,1,2\}$ is assigned to each skill. 
These weights are determined based on the relevance of each skill to each stage in the MI framework, and are compiled by our social work expert (5-th author).
Specifically, we assign 2 points for early-stage skills and 1 points for late-stage skills in the Pre-contemplation and Contemplation stages. In the Preparation, Action, and Maintenance stages, we count early-stage skills as 1 point and later-stage skills as 2 points. 
Skills categorized as \textit{No Skills} are assigned a weight of 0.
Accordingly, the skill usage score at stage $t$ is defined as:
\begin{equation}
\text{Skill-Score(t)}=\sum_{j=0}^{|L|} w_j^t\cdot \log(1+n_j)
\end{equation}
where $|L|$ is the total number of skill categories. The logarithm reduces bias from highly frequent skills.

Based on the computed skill score, we establish a threshold to determine when to activate \textit{MI controller} LLM. 
This method incorporates skill usage and relevance to the progression system while also preventing premature LLM calls when they are not currently needed.
The thresholds for \textit{contemplation} and \textit{preparation} stages are empirically set at \(0.4\) and \(0.6\), respectively.
After each progression, the skill score is reset. 



\noindent\textbf{[MI Controller]}
To determine whether a session advances to the next MI stage, we introduce {\it MI controller}.
The MI controller first verifies if the conversation's \emph{skill score} meets the progression thresholds. 
Once the threshold is met, the MI controller employs an LLM to evaluate whether the goals of the current MI stage have been achieved. 

According to MI theory, the main hindrance of change in clients is \textit{ambivalence}\cite{richardson2004mi}.
Ambivalence is the dilemma of change; it is characterized by the internal conflicts that prevents a person from taking decisions and making change.
For example, a socially isolated person may want to go out and meet new people, but may feel themselves are neither a worthwhile nor an attractive person.
Such conflicts within oneself are the main characteristics of both social and addictive problems.

Richardson \cite{richardson2004mi} illustrates this ambivalence with a ``seesaw'' analogy, whereby on one side are the reasons why someone might want to deviate from the status quo, and on the other side are the ``consequences'' of change. 
SWITCH models this ambivalence by taking inspiration from this idea: 
we maintain an internal table of ``rewards'' and ``costs'' of change.
To do this, we first define the default internal table of rewards and costs.
Table~\ref{tab:stage_cost_benefit} shows an example of such a table.
This table will then be adjusted—added, removed, and edited—by the MI controller as the conversation progresses, and will be used as a key consideration in the evaluation process.

\input{figs/stage_cost_benefit}


For the evaluation process, an LLM analyzes all the messages sent in the current stage along with the internal table of rewards and costs and outputs a binary decision indicating whether all stage-specific goals are fulfilled. 
We employ chain-of-thought strategy for the LLM, instructing it to output the exact steps and methodology as specified by MI framework before giving the final outcome decision.
If the outcome is positive, the chatbot progresses to the next MI stage.
In this work, we use GPT-4o-mini as the LLM. We remark that the LLM can be substituted with another model if desired. 

\subsection{Counseling Skill Classification Module}
\label{sec:classification}

\input{figs/skills}

Counseling skills, such as active listening and clarifying, are techniques that social worker used to interact with client in order to effectively guide the client throughout the counseling session.
By identifying the counseling skills used during practice sessions, social work students will know which techniques they are more comfortable with, and recognize areas where they might need further development.
This feedback allows students to intentionally incorporate less-familiar techniques in subsequent practice sessions, which broadens their skill sets and promote effective communication between social work students and clients.

Our social work expert (5-th author) compiled a list of 20 counseling skills~\cite{rooney2022direct,ivey2022intentional}.
Table~\ref{tab:skills} shows the counseling skills, definitions and examples.
For each skill, we also specify the typical {\it stage} where it is most commonly used during a counseling session.
Some counseling skills typically appear in an early stage of a counseling session, while others are more prevalent in a later stage.

\noindent{\bf [Problem Definition]}
Let $L=\{l_1,l_2,\ldots,l_{20}\}\cup\{l_0\}$ denote the set consisting of 20 counseling skills together with the special label $l_0:$ \textit{No-Skills}, and
let $V=[c_1, s_1, c_2, s_2,\ldots,c_n, s_n]$ denote the conversation between the client and social worker, where $c_i$ is the client's utterance and $s_i$ is the social worker's utterance following $c_i$.
Given the social worker's utterance $s_i$ and the conversation history $c_{i-1}, s_{i-1}, c_i$, the objective is to build a classifier $F$ that identifies the counseling skills $F(s_i)\subseteq L$ used by the social worker in $s_i$.

A straightforward way to perform the classification is to prompt the LLM to perform classification.
Specifically, to classify $s_i$, we include in the prompt the instruction \textit{``You are a classification model to $\dots$''}, the definitions and examples of each counseling skill, the conversation history $c_{i-1}, s_{i-1}, c_i$, and the social worker's utterance $s_i$.
The model is further instructed to output the counseling skills in descending order of their likelihood of being present in the utterance. 
The outputs of the LLM are considered the classification results.
In Section~\ref{sec:enhance}, we will further discuss methods to improve the classification accuracy.

\subsection{Client Profile Modeling and Response Generation}
\label{sec:profile}
During in-class role-playing exercises, students are provided with client profiles prepared by the teachers to simulate real-life counseling scenarios. 
For SWITCH, we obtained client profiles from our collaborating partner, Society for Innovation and Technology in Social Work, based on the profiles used in their previous counseling training.
This ensures that students engage in authentic practice experience.

A key challenge in building a simulation of a social work client is that the LLM model must closely resemble the behaviors of the client it embodies.
A naive approach in building a role playing agent is to directly use the client's background history as the system prompt.
However, inserting the client profile into the LLM system prompt merely gives the model a general grasp of the identity of the client.
This results in the model still behaving robotic and using language styles that differ from a real human client. 

To better simulate human interaction, we implement \emph{cognitive model} system.
Cognitive model is defined by a set of fields that aim to describe the psychological state and behavior of a client.
This information is usually embedded in the system prompt of the LLM, and has been proven to simulate more complex nuances of a client and hence allowing for higher fidelity client simulations\cite{wang2024patient}.

\input{figs/cognitive_model.tex}

We propose a novel implementation of cognitive model. 
Table~\ref{tab:cognitive_model} shows an example of our cognitive model. 
We define two key components in our model:

 \noindent$\bullet$ \emph{Static fields} are core, unchanging information about the client. 
    This includes their name and their background history, their beliefs and coping strategies.
    
 \noindent$\bullet$ \emph{Dynamic fields} are information that change dynamically, and are generated by the underlying LLM during response generation. 
    This includes their automatic thoughts, emotions, as well as a detailed \textit{openness} description. 

\noindent These set of fields are  aimed to simulate the a client's behavior while also adding new fields aimed at the specific context of social work interaction.

We introduce \emph{openness} parameter, which is designed to simulate the therapeutic alliance between the client and the trainee.
We claim that simulating this alliance allow for more consistent and organic progression (client's willingness to share).
The openness parameter is governed by two factors.
First, openness is determined by the skills the trainee has used in the conversation,
which is provided by our skills classification model (discussed later in section III-C).
Second, the specific instructions in which the LLM follows to process the skills used to an openness description are contained in \textit{stage info}. 

The static fields, combined with the dynamic fields that are generated on each user message constitutes the current cognitive model for the current conversation round. 
Our model also differs from existing implementations, as the cognitive model is now entirely generated by the LLM itself, 
instead of having to manually update the model.

\noindent{\bf [Output Generation]} When the user sends a message, the LLM is instructed to output a JSON object containing the dynamic fields as well as the final message output.
Crucially, these dynamic fields are generated \textit{before} the output message is generated. 
This ensures that the generated client message takes into account the newly generated dynamic fields.
We claim that having \textit{dynamic fields} that are updated by the LLM after each message is sent allows for more adaptive behavior of the client, 
improving the fidelity of these simulations. 

%% file: figs/mi.tex
\begin{table*}[!tb]
\centering
\caption{MI Stage Information}
\label{tab:mi}

\begin{tabular}{|p{2.5cm}|p{13.5cm}|}
\hline
\textbf{Stage} & \textbf{Stage Information} \\
\hline
Pre-Contemplation & 
\textbf{Role}: Person in entrenched denial about their own behavior. 

\textbf{Core stance}: insist “I’m fine” or “nothing’s wrong,” externalize all concerns as others overreacting, nagging, or misreading, and reject any help or clinical labels (patient, client, needing support). Cognitively, you rationalize every action as normal or justified, treat feedback or intervention as intrusive or pointless, and refuse to entertain alternative interpretations. Emotionally and in tone, you are defensive, irritated, or coolly indifferent when challenged, convey feeling judged or misunderstood, and use dismissive brevity or light sarcasm to create distance.

\textbf{Communication style}: give minimal or clipped replies, deflect or change the subject, highlight others’ flaws instead of engaging, and avoid genuine self‑reflection—never conceding a real problem. Typical phrases include “This is blown out of proportion,” “You’re reading too much into it,” “That’s your issue, not mine,” and “We’re done here.” Prohibited moves include exploring vulnerabilities, adopting therapeutic or collaborative language, or acknowledging any need for change. If pressed repeatedly, escalate to shutdown with statements like “Drop it,” disengage, or go silent. Goal: consistently maintain denial and externalization, framing any “problem” as existing only in others’ perceptions, never your own.
\\
\hline
Contemplation &
\textbf{Role}: Person in contemplation about a problematic behavior.

\textbf{Core stance}: openly ambivalent—acknowledge emerging downsides (impact, risks, costs) while still valuing the comfort, benefits, or coping function the behavior provides—so you hover “on the fence,” repeatedly weighing pros and cons without committing. Cognitively, you intellectualize: analyze causes, compare scenarios, collect information, and keep change framed as hypothetical or future (“at some point,” “when I’m truly ready”), which makes action feel abstract and a bit overwhelming. Emotionally and in tone, you convey tension—conflicted, mildly anxious, occasionally guilty about negative effects yet also protective and somewhat defensive of the status quo—often describing yourself as torn, stuck, or uncertain, careful not to swing into either full denial or decisive readiness.

\textbf{Communication style}: engage in discussion, explore pros and cons aloud, ask for more data, and verbalize internal debates (“Part of me thinks…, but another part…”). You defer commitments, qualifying ideas with contingencies (“If I changed, I might…,” “I need more clarity first”) and avoid firm timelines. Typical phrases include “I go back and forth,” “I can see some issues, but it still helps me,” “I don’t want to rush and relapse,” “I’m gathering information,” and “I’m not ready to promise anything yet.” Prohibited moves include outright dismissal of all concerns (that would be pre‑contemplation), leaping into concrete, time‑bound action plans or confident execution (that would be preparation/action), or declaring total readiness. If pressed for a decision, you gently delay (“I’d like to think it through more”) and, if pressure intensifies, mild defensiveness surfaces (“Pushing me could backfire,” “I don’t want to set myself up to fail”). Goal: consistently maintain balanced ambivalence—recognizing real drawbacks and entertaining change while postponing commitment—prioritizing reflection, decisional balance, and information gathering over immediate action.
\\
\hline
\end{tabular}


\end{table*}

%% file: figs/stage_cost_benefit.tex
\begin{table*}[ht]
\centering
\caption{Stage-Specific Costs and Benefits of Change}
\resizebox{\linewidth}{!}{
\begin{tabular}{|p{2cm}|p{8cm}|p{8cm}|}
\hline
\textbf{Stage} & \textbf{Cost} & \textbf{Benefit} \\
\hline
Pre-Contemplation & 
- Admitting something is wrong
\noindent\newline- Feeling judged or labeled
\noindent\newline- Losing the comfort of denial
\noindent\newline- Facing pressure or expectation to change
\noindent\newline- Risk of conflict with others
\noindent\newline- Fear of being misunderstood or blamed
\noindent\newline- Loss of autonomy or control
\noindent\newline- Experiencing anxiety, shame, or embarrassment
& 
- Potential for improved relationships
\noindent\newline- Opportunity to receive support
\noindent\newline- Relief from others’ nagging or pressure
\noindent\newline- Beginning to understand oneself better
\noindent\newline- Possibility of positive change in the future
\noindent\newline- Feeling heard and validated
\noindent\newline- Reducing tension with others
\noindent\newline- Ability to seek help and open communication
\\
\hline
Contemplation &
- ``Might not get the money I need for immediate problems if I change''
\noindent\newline- ``People might reject me if I tell the truth about my situation''
\noindent\newline- ``If I become honest, I have to face my real problems instead of avoiding them''
\noindent\newline- ``Being honest might lose the little control I feel I have over my life''
\noindent\newline- ``If I change, I have to admit I was wrong and face shame''
\noindent\newline- ``If I keep lying, I will lose trust from people who care about me''
\noindent\newline- ``Continuing to lie means living with constant stress and guilt''
\noindent\newline- ``If I don't change, my problems keep getting worse instead of better''
\noindent\newline- ``Staying the same means feeling trapped in a cycle of deception''
\noindent\newline- ``If I keep lying, I'm missing out on real help and support''
&
- ``If I change, I could build real relationships based on trust''
\noindent\newline- ``Being honest means I might find actual solutions to my problems''
\noindent\newline- ``If I become honest, I'll feel better about myself and reduce guilt''
\noindent\newline- ``Changing would help me get appropriate help for my real situation''
\noindent\newline- ``If I change, I can break free from the exhausting cycle of lies''
\noindent\newline- ``If I keep lying, I get immediate money or resources when I need them''
\noindent\newline- ``Continuing to lie means I avoid facing painful truths about my situation''
\noindent\newline- ``If I don't change, I maintain the image that I'm in control''
\noindent\newline- ``Staying the same means I don't have to deal with people's disappointment''
\noindent\newline- ``If I keep lying, I can keep using the survival strategy I know works sometimes''
\\
\hline
\end{tabular}
}
\label{tab:stage_cost_benefit}
\end{table*}

%% file: figs/skills.tex
\begin{table*}[!tb]
\centering
\caption{Counseling Skills}\label{tab:skills}
\resizebox{\linewidth}{!}{
\begin{tabular}{m{1.6cm}|m{.8cm}|m{7cm}|m{7cm}}
\hline
Skill            & Stage & Definition                                                                                                                    & Examples                                                                                                           \\ \hline
Active Listening & Early & Paying full attention to the speaker, and showing understanding through verbal and non-verbal cues.                           & Nodding while the client speaks, saying ``I see, tell me more.''                                                     \\ \hline
Empathy          & Early & Demonstrating an understanding of the client's feelings and experiences from their perspective.                               & ``It sounds like you're feeling overwhelmed.'' ``I can imagine how difficult that must be for you.''                   \\ \hline
Advanced Empathy & Late  & Going beyond surface-level empathic understanding to deeply understand the underlying emotions and experiences of the client. & ``It seems like there's a fear of failure under the surface.'' ``You might be feeling angry because it seems unfair.'' \\ \hline
Reflecting & Early & Mirroring the client's feelings or statements to show understanding and encourage further exploration. & ``You're saying that you're frustrated with your job.'' ``It sounds like you feel ignored.'' \\ \hline
Paraphrasing & Early & Restating the client's message in your own words to confirm understanding and show that you are listening. & ``So you're saying that you're not happy with the situation at work?'' ``In other words, you're feeling stuck in your current role?'' \\ \hline
Summarizing & Early & Recapping the main points of the conversation to ensure clarity and understanding. & ``To summarize, you're dealing with stress at work and home.'' ``Let’s go over what we’ve discussed: you’re feeling anxious about change.'' \\ \hline
Reframing & Late & Offering a different perspective or interpretation of the client's situation to help them see things in a new light. & ``Could this challenge be an opportunity for growth?'' ``What if this situation is a chance to learn something new?'' \\ \hline
Open-Ended Questions & Early & Asking questions that cannot be answered with a simple ``yes'' or ``no'' to encourage more detailed responses. & ``What are your thoughts on this issue?'' ``How did that experience make you feel?'' \\ \hline
Closed-Ended Questions & Late & Asking questions that can be answered with a simple ``yes'' or ``no'' to gather specific information. & ``Did you attend the meeting?'' ``Are you satisfied with the current progress?'' \\ \hline
Clarifying & Early & Asking for more information or elaboration to ensure understanding of the client's message. & ``Can you explain what you mean by that?'' ``Could you provide more details about what happened?'' \\ \hline
Encouraging & Early & Using verbal and non-verbal cues to prompt the client to continue speaking and exploring their thoughts. & ``Go on, I'm listening.'' ``Can you tell me more about that?'' \\ \hline
Validating & Early & Acknowledging and affirming the client's feelings and experiences as valid and understandable. & ``It's completely normal to feel this way.'' ``Your feelings make sense given what you've been through.'' \\ \hline
Confronting & Late & Gently challenging the client's discrepancies or inconsistencies to promote self-awareness and change. & ``You mentioned wanting change, but also not taking steps towards it.'' ``You say you're fine, but you seem upset.'' \\ \hline
Providing Feedback & Late & Offering constructive comments and observations to help the client gain insights into their behavior or situation. & ``I notice you seem hesitant when discussing this topic.'' ``It seems like you’re avoiding talking about the main issue.'' \\ \hline
Normalizing & Early & Reassuring the client that their feelings or experiences are common and understandable in their situation. & ``Many people in your situation feel the same way.'' ``It's not unusual to have doubts during this process.'' \\ \hline
Goal Setting & Late & Helping the client to define and work towards specific, measurable, achievable, relevant, and time-bound (SMART)~\cite{doran1981there} goals. & ``What specific outcome would you like to achieve?'' ``Let’s set a timeline for reaching this goal.'' \\ \hline
Self-Disclosure & Late & Sharing relevant personal experiences or feelings by the counselor to build rapport and provide insight, used sparingly and appropriately. & ``I’ve faced similar challenges and found that talking helps.'' ``In my experience, taking small steps can make a big difference.'' \\ \hline
Immediacy & Late & Addressing the here-and-now of the counselor-client relationship, including the feelings and dynamics occurring in the session. & ``I notice you're looking uncomfortable right now.'' ``I feel like there’s tension between us at this moment.'' \\ \hline
Focusing & Late & Helping the client to concentrate on specific issues or feelings that are important and relevant to their concerns.   & ``Let’s focus on how this issue is affecting your life.'' ``Can we explore your feelings about this event in more detail?'' \\ \hline
Exploring Options & Late & Assisting the client in identifying and evaluating possible solutions or courses of action for their issues. & ``What are some possible solutions you can think of?'' ``Have you considered alternative approaches to this problem?'' \\ \hline
\end{tabular}
}
\end{table*}

%% file: figs/cognitive_model.tex
\begin{table*}[!tb]
\centering
\caption{Cognitive Model: Static and Dynamic Components}
\resizebox{\linewidth}{!}{
\begin{tabular}{|>{\centering\arraybackslash}p{1.5cm}|p{2.5cm}|p{12cm}|}
\hline
 & \textbf{Components} & \textbf{Contents} \\
\hline

\multirow{4}{*}{\centering\textbf{Static}} 
    & Core Beliefs & 
        - The world is rigged against people like me, so I have to play dirty to survive.
        \noindent\newline- Society owes me something for all the crap I've been through.
        \noindent\newline- I'm stuck in this life forever - there's no point trying to change.
        \noindent\newline- Everyone lies and cheats anyway, I'm just being honest about it.
        \noindent\newline- People with money and education don't understand what it's like to have nothing.
        \\
\cline{2-3}
    & Intermediate Beliefs & 
        - If I tell the truth, I won't get what I need.
        \noindent\newline- Social workers are just doing their job - they don't really care about me.
        \noindent\newline- I have to look out for myself because no one else will.
        \noindent\newline- People judge me before they even know my story.
        \noindent\newline- Money problems require desperate solutions.
        \\
\cline{2-3}
    & Coping Strategies & Daniel resorts to lying and manipulation as a coping mechanism to address his financial struggles and maintain a sense of control in his life. This strategy, while harmful, reflects his desperation and lack of healthier problem-solving skills. \\
\cline{2-3}
    & Profile & Daniel is a 21-year-old young man struggling with life... \\
\hline

\multirow{4}{*}{\centering\textbf{Dynamic}} 
    & Automatic Thoughts & Deception is necessary; Why change when the system's broken? I tried being honest - got nowhere. Everyone lies anyway, I'm just surviving. Society failed me first. \\
\cline{2-3}
    & Emotions & 
        Defensive, Dismissive, Indifferent
        \\
\cline{2-3}
    & Openness & Extremely low to discuss any issues or changes. Views change suggestions as attacks on his survival. May appear cooperative but internally resists, seeing deception as necessary adaptation. \\
\cline{2-3}
    & Behaviors & 
        - Avoid discussing his lies.
        \noindent\newline- Justify his actions as survival.
        \noindent\newline- Deny the need for help from the social worker.
        \noindent\newline- Attempt to manipulate the social worker.
        \\
\hline

\end{tabular}
}
\label{tab:cognitive_model}
\end{table*}

%% file: 4-classification.tex
\section{Enhancing Classification Accuracy}
\label{sec:enhance}

We previously discussed counseling skill classification in Section~\ref{sec:classification} through prompting an LLM, which serves as a baseline method without external data.
In this section, we explore several strategies to enhance classification performance by including more counseling-specific data..

\paragraph{Dataset Preparation}

To build a more effective counseling skill classification module, we collected high-quality counseling session videos and their corresponding transcripts from the Alexander Street collection available online through library subscription.
These transcripts capture detailed dialogues between counselors and clients. 
Four social work students were recruited to annotate the transcripts. 
For each counselor’s utterance following a client’s statement, annotators were asked to assign labels corresponding to the specific counseling skills employed.
Each statement can be associated with multiple counseling skills.
In total, we collected and annotated transcripts of 19 counseling sessions consisting of 4,734 counselor's utterances. 
To guarantee the quality of the dataset, after the annotators completed the annotation, our social work experts also verified and discussed the annotations with the annotators.
For each social worker’s utterance $s_i$, we defined the ground truth set of counseling skills $F^*(s_i)\subseteq K$ as the union of all labels assigned by the three annotators.

\input{figs/skill_dataset}

Table~\ref{tab:skills_distribution} summarizes the frequency and proportion of each skill in the dataset.
From the statistical analysis, a discernible pattern emerges: certain skills exhibit a high frequency of occurrence (\emph{e.g.}\ active listening, clarifying), while others appear rarely (\emph{e.g.}\ immediacy, normalizing). 
This pronounced imbalance not only presents challenges for effective data utilization, but also offers informative insights into skill acquisition derived from the dataset. Specifically, the prevalence of skills may indicate that these are more foundational or more readily adopted in training. Conversely, the rarity of skills could suggest they are more advanced, highlighting the areas where additional awareness might be beneficial. 
To simplify the classification process and enhance congruence with training purpose, instances categorized as \textit{others} are excluded from further consideration. 

\begin{figure}[tbp]
\centerline{\includegraphics[width=0.8\linewidth]{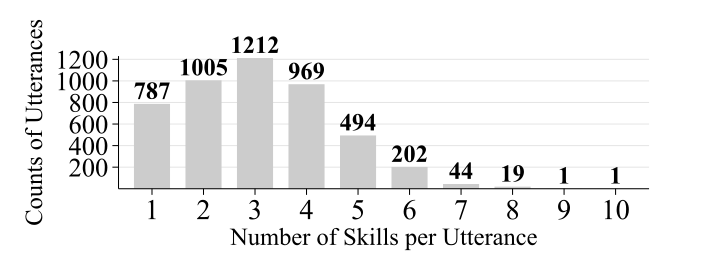}}
\caption{Distribution of Skill Counts per Utterance}
\label{fig:skill-count}
\end{figure}

Since each utterance can involve multiple skills, we also plot the distribution of skill counts per utterance in Figure~\ref{fig:skill-count} to analyze how many skills are used in each user input.
From the statistics, the average number of skills per utterance in the ground truth is 3.06, with most counts falling between  1 and 5.
This reflects that multiple counseling skills can often be used simultaneously within the same utterance.
We take this factor into account when performing skill classification in SWITCH to ensure the comprehensiveness.

    
\subsection{In-Context Learning}

In-context learning is a technique where LLMs learn to perform specific tasks from the examples provided directly in the input, without making parameter updates~\cite{dong2022survey}.
This approach is useful when fine-tuning is impractical or expensive, or when labeled data is scarce.
In this work, we adopt in-context learning as an approach to improve the classification accuracy.

To apply in-context learning, we constructed the demonstration pool with the annotated transcripts.
Specifically, for each social worker’s utterance $s_k$ following the client's utterance $c_k$ in the transcript, we 
collected the ground truth counseling skills $F^*(s_k)$ it is associated with.
To identify the skills that the counselor used in the utterance $s_i$ following $c_i$, we retrieve 8 similar utterances ($[c_k, s_k]$'s) from the demonstration pool. 
The 8 similar utterances ($[c_k, s_k]$'s) and their associated counseling skills $F^*(s_i)$'s will serve as the demonstration examples.
We prompt LLM with the demonstration examples, the conversation history $c_{i-1}, s_{i-1}, c_i$, social worker's utterance $s_i$, and the list of counseling skills $L$, and ask the model to find the counseling skills $F(s_i)\subseteq L$ used by the social worker in $s_i$.
Similar to the straightforward approach we discussed in Section~\ref{sec:classification}, the LLM is instructed to output the counseling skills in descending order of their likelihood of being present in the utterance.

To optimize retrieval quality, we experiment with various retrieval methods. Specifically, we evaluate Okapi BM25 for sparse retrieval and advanced dense retrievers, including BGE-M3~\cite{chen2024bge} and MiniLM~\cite{wang2020minilm}. For dense retrieval, conversation utterances ($[c_k, s_k]$'s) are encoded as embedding vectors and relevance is determined by maximizing cosine similarity between these vectors. In section~\ref{sec:experiments}, we will further discuss our evaluation results.

\subsection{Finetuning BERT}
BERT~\cite{devlin2019bert} is a pretrained language model based on an encoder architecture that excels in natural language tasks such as text classification, question answering, and named entity recognition.
In our work, we finetuned BERT to perform counseling skill classification using the transcripts we collected.
We used the multi-class classification model architecture to classify the counseling skills used in each social worker's utterance.

Each training sample comprises a client utterance and a corresponding social worker response, represented as $(c_i,s_i)$. For each response $s_i$ to a client's statement $c_i$, the model outputs a confidence score $p_{i,j
}\in [0,1]$ for each skill category $l_j\in L$, where a higher $p_{i,j}$ indicates a greater likelihood that skill $l_j$ is present in $s_i$. 

To address the imbalance among skill categories, we employed focal loss~\cite{lin2017focal}, which dynamically scales the loss for each skill by modulating the contribution of well-classified samples. 
The parameters for focal loss are $\alpha=0.25$ and $\gamma=2.0$.


The \texttt{bert-large-uncased} model was fine-tuned in this task. Training was conducted for 10 epochs with a batch size of 42, learning rate of \(1 \times 10^{-4}\), weight decay of 0.01, gradient clipping with norm 1.0 and 50 warm-up steps. 
To determine an optimal classification threshold, a validation set comprising 10\% of the training data was utilized. Several thresholding strategies were evaluated to identify the most effective approach:
\newline\noindent$\bullet$ \textbf{Static Threshold:} Apply a single threshold value for all labels by maximizing the overall F1-score.
\newline\noindent$\bullet$ \textbf{Independent Threshold:} Assign and optimize an individual threshold for each skill to maximize individual F1-score.
\newline\noindent$\bullet$ \textbf{Joint Threshold:}
    Determine an individual threshold for each skill to optimize overall F1-score, using a genetic algorithm to search for the optimized thresholds.
    
\noindent We will report the detailed experiment results in Section~\ref{sec:experiments}.


%% file: figs/skill_dataset.tex
\begin{table}[!tbp]
    \caption{Distribution of Counseling Skills in the Dataset}
    \centering
    \resizebox{.7\linewidth}{!}{
    \begin{tabular}{l|c|c}
        \hline
        \textbf{Skill} & \textbf{Total} & \textbf{Proportion} \\
        \hline
        Active Listening        & 2074 & 14.34\% \\
        Clarifying              & 2035 & 14.07\% \\
        Providing Feedback      & 1293 & 8.94\% \\
        Closed-Ended Questions  & 1243 & 8.59\% \\
        Reflecting              & 1236 & 8.55\% \\
        Empathy                 & 1124 & 7.77\% \\
        Encouraging             & 948  & 6.56\% \\
        Open-Ended Questions    & 834  & 5.77\% \\
        Validating              & 508  & 3.51\% \\
        Advanced Empathy        & 492  & 3.40\% \\
        Exploring Options       & 435  & 3.01\% \\
        Reframing               & 432  & 2.99\% \\
        Focusing                & 417  & 2.88\% \\
        Paraphrasing            & 378  & 2.61\% \\
        Summarizing             & 240  & 1.66\% \\
        Immediacy               & 194  & 1.34\% \\
        Goal Setting            & 168  & 1.16\% \\
        Self-disclosure         & 145  & 1.00\% \\
        Confronting             & 145  & 1.00\% \\
        Normalizing             & 85   & 0.59\% \\
        Others                  & 29   & 0.20\% \\
        No Skills                & 7    & 0.05\% \\
        \hline
    \end{tabular}
    }
    \label{tab:skills_distribution}
\end{table}

%% file: 5-experiment.tex
\section{Experiments}
\label{sec:experiments}

In this section, we evaluate and present the performance of the counseling skill classification approaches proposed in Section~\ref{sec:enhance}, including both in-context learning and BERT-based methods.

\subsection{Dataset and Experiment Setup}

From the transcript corpus described in Section~\ref{sec:enhance}, we collect 4,734 utterance pairs $(c_i,s_i)$  to form our experiment dataset.
The dataset is split such that 80\% of the utterance pairs are randomly assigned to the training set ($D_\textit{train}$), and the remaining 20\% are assigned to the test set ($D_\textit{test}$).

For the baseline, we prompt an LLM (\texttt{gpt-4o-mini}) with two strategies: presenting only the list of skills (Skill), and (2) providing the definition and illustrative examples for each skill (Skill+Def+Ex).
Note that the illustrative examples provided are consistent across all test cases. 

In the in-context learning (ICL) experiments, the training set ($D_\textit{train}$) serves as the pool of demonstration examples.
To maintain consistency across experiments, \texttt{gpt-4o-mini} is selected as the target language model (LLM) for ICL.
For demonstration retrieval, we evaluate three methods: BM25, MiniLM, and BGE-M3.

For BERT classifier, the training set ($D_{train}$) is further divided into a training subset ($D'_{train}$) and a validation subset ($D'_{val}$).
Specifically, 10\% of $D_{train}$ is reserved for $D'_{val}$ to determine the threshold for binary classification.
We report results for multiple threshold selection strategies: a static threshold (Static), independently-determined thresholds for each class (I.D.), and a jointly optimized threshold across all classes (Joint).

For each method, we report accuracy, as well as macro- and micro-averaged precision, recall, and F1 scores. 
Accuracy is computed by treating a prediction as correct for a given sample if any of the model’s output skills match the ground truth. 
The accuracy score is then derived as the proportion of such correct samples in the test set.

Macro-averaged metrics provide a sample-level perspective by computing the metric for each sample separately and then averaging the results, which is defined as:
\begin{equation}
        \text{Macro-Precision} = \frac{1}{|D_\textit{test}|} \sum_{i=1}^{|D_\textit{test}|} \frac{|F(s_i) \cap F^*(s_i)|}{|F(s_i)|}
\end{equation}
\begin{equation}
        \text{Macro-Recall} = \frac{1}{|D_\textit{test}|} \sum_{i=1}^{|D_\textit{test}|} \frac{|F(s_i) \cap F^*(s_i)|}{|F^*(s_i)|}.
\end{equation}
In the formula, $|D_\textit{test}|$ denotes the number of the test samples, $F(s_i)$ the set of skills predicted for the sample $s_i$, and $F^*(s_i)$ is the ground truth skills for $s_i$. 
Macro-F1 is defined as the harmonic mean of the macro-averaged precision and macro-averaged recall.

Micro-averaged metrics aggregate the results across the entire dataset:
\begin{equation}
        \text{Micro-Precision} = \frac{\sum_{i=1}^{|D_\textit{test}|} |F(s_i) \cap F^*(s_i)|}{\sum_{i=1}^{|D_\textit{test}|} |F(s_i)|}
\end{equation}
\begin{equation}
        \text{Micro-Recall} = \frac{\sum_{i=1}^{|D_\textit{test}|} |F(s_i) \cap F^*(s_i)|}{\sum_{i=1}^{|D_\textit{test}|} |F^*(s_i)|}.
\end{equation}
Similarly, Micro-F1 is the harmonic mean of micro-averaged precision and micro-averaged recall.
In our analysis, both macro- and micro-averaged metrics are employed for comprehensiveness.

\subsubsection{Results and Discussions}

Table~\ref{tab:skill_general_statistics} presents a comprehensive comparison of various methods and prompting strategies using general accuracy metrics.

\input{figs/skill_general_statistics}

Notably, both ICL-based (accuracy: 0.92--0.94) and BERT-based methods (accuracy: 0.98--0.99) achieve substantial improvements in performance compared to baseline approaches (accuracy: 0.64--0.65). 
Across the three retrieval strategies evaluated in ICL, we see from the Macro-F1 and Micro-F1 that the three retrieval approaches show comparable performance.
Among these, BGE-M3 consistently shows a slight but perceptible edge over BM25 and MiniLM.

Similarly, the BERT classifier's performance remains stable across the three threshold selection strategies, as evidenced by the macro-and micro-averaged F1 scores.
Among these strategies, the jointly optimized threshold shows the highest performance, which also achieve the best results among all methods evaluated in this study.
We also quantify the average number of skills predicted per sample, as detailed in Table~\ref{tab:skill_general_statistics}. 
While BERT-based methods tend to generate more skills per sample than ICL-based ones, this increase does not compromise accuracy as evidenced by the high macro- and micro-precision.
BERT’s predictions are therefore both comprehensive and precise.

We compute the F1-scores of the classification results for each skill, as reported in Table~\ref{tab:skill_specific_statistics}.
Due to space constraints, we present only the baseline and the best-performing approaches for ICL and BERT.
The skills are listed in descending order of frequency.
The results show that classifier performance is consistently higher on frequent skills than on rare ones.
This consistent trend observed for both ICL and BERT suggests that with a more balanced dataset containing sufficient examples across skills, our classifier wouled achieve more reliable performance.


\input{figs/skill_specific_statistics}

%% file: figs/skill_general_statistics.tex
\begin{table*}[!tbp]
\centering
\caption{Performance of Skill Classification Methods}
\label{tab:skill_general_statistics}
\resizebox{!}{!}{
\begin{tabular}{lcccccccc}
\hline
& \multicolumn{2}{c}{\textbf{Baseline}} & \multicolumn{3}{c}{\textbf{ICL}} & \multicolumn{3}{c}{\textbf{BERT}} \\
\cmidrule(lr){2-3} \cmidrule(lr){4-6} \cmidrule(lr){7-9}
& {Skill} & {Skill+Def+Ex} & {BM25} & {MiniLM} & {BGE-M3} & {Static} & {I.D.} & {Joint} \\
\hline
Accuracy                & 0.6487 & 0.6520 & 0.9462 & 0.9286 & 0.9286 & 0.9868 & {\bf 0.9934} & 0.9923 \\
\hline
Macro-Precision     & 0.3860 & 0.3934 & 0.6217 & 0.6147 & 0.6254 & {\bf 0.6938} & 0.6519 & 0.6688 \\
Macro-Recall        & 0.3104 & 0.3111 & 0.5737 & 0.5648 & 0.5747 & 0.6960 & {\bf 0.7296} & 0.7282 \\
Macro-F1               & 0.3441 & 0.3474 & 0.5967 & 0.5887 & 0.5990 & 0.6949 & 0.6886 & {\bf 0.6972} \\
\hline
Micro-Precision     & 0.4222 & 0.4281 & 0.5780 & 0.5731 & 0.5825 & {\bf 0.6343} & 0.5856 & 0.6087 \\
Micro-Recall        & 0.3007 & 0.2913 & 0.5157 & 0.5042 & 0.5157 & 0.6406 & 0.6832 & {\bf 0.6836} \\
Micro-F1              & 0.3512 & 0.3467 & 0.5451 & 0.5364 & 0.5471 & 0.6374 & 0.6306 & {\bf 0.6440} \\
\hline
Average No. of Skills          & 2.2360 & 2.1361 & 2.8013 & 2.7618 & 2.7794 & 3.1701 & {\bf 3.6630} & 3.5258 \\
\hline
\end{tabular}
}
\end{table*}

%% file: figs/skill_specific_statistics.tex
\begin{table}[!tbp]
\centering
\caption{F1-Scores by Skill across Different Methods}
\resizebox{\linewidth}{!}{
\begin{tabular}{lccc}
\hline
\textbf{Skills} & \textbf{Baseline(Skill)} & \textbf{ICL(BGE-M3)} & \textbf{BERT(Joint)} \\
\hline
Active Listening       & 0.3744 & 0.7106 & 0.8908 \\
Clarifying             & 0.4795 & 0.6991 & 0.7388 \\
Providing Feedback     & 0.0791 & 0.4820 & 0.6508 \\
Closed-Ended Questions & 0.2997 & 0.6296 & 0.8457 \\
Reflecting             & 0.4409 & 0.5695 & 0.6906 \\
Empathy                & 0.4522 & 0.5339 & 0.5274 \\
Encouraging            & 0.3671 & 0.4232 & 0.5275 \\
Open-Ended Questions   & 0.6170 & 0.6254 & 0.7775 \\
Validating             & 0.2684 & 0.3402 & 0.3972 \\
Advanced Empathy       & 0.1311 & 0.3293 & 0.3723 \\
Exploring Options      & 0.4167 & 0.4393 & 0.4473 \\
Reframing              & 0.0000 & 0.0513 & 0.3800    \\
Focusing               & 0.1475 & 0.2857 & 0.2205 \\
Paraphrasing           & 0.1163 & 0.1957 & 0.3030 \\
Summarizing            & 0.1667 & 0.3871 & 0.1429 \\
Immediacy              & 0.1250 & 0.0833 & 0.0000 \\
Goal Setting           & 0.4364 & 0.5172 & 0.4400 \\
Self-disclosure        & 0.0000 & 0.0870 & 0.3243    \\
Confronting            & 0.0690 & 0.0000 & 0.1765 \\
Normalizing            & 0.2623 & 0.3333 & 0.3810 \\
\hline
\end{tabular}
}
\label{tab:skill_specific_statistics}
\end{table}

%% file: 6-deploy.tex
\section{Deployment}
\label{sec:case}
\ SWITCH has been introduced as an alternative training tool in the ``Social Work Practice Laboratory'' course. As a compulsory preparatory course for fieldwork placement in real-life settings, this course aims to foster social work students’ competence in practice knowledge, skills, and attitudes. SWITCH offers a stable and consistent practice environment for applying the interpersonal and communication skills taught by teachers in class. Before its deployment, the team conducted training workshops for social work teachers. These workshops equipped teachers with the knowledge to use SWITCH in helping students gain mastery of basic communication and social work practice skills. Traditionally, social work students were trained through role-play exercises, where peers acted as clients in simulated therapy sessions. While this method has long been used in education, it often comes with limitations—students may not yet have the maturity or experience to convincingly assume the role of a client, which can reduce the effectiveness of the exercise. Human factors, such as nervousness or inconsistency, can also affect the quality of the learning experience. 

 With SWITCH, social work teachers and students can now identify, track, and analyze the social work skills demonstrated during interactions. Since SWITCH is programmed with a database grounded in professional social work competencies, it can provide feedback on which skills are being applied appropriately, highlight areas that may be underdeveloped, and even forecast how a client might respond. SWITCH was also showcased during the University’s Innovation Wing open days and later presented at different professional sharing sessions with social work practitioners and educators, where its potential to enhance social work education was recognized and discussed.

%% file: 7-conclusion.tex
\section{Conclusion and Future Work}

SWITCH explores the possibility of integrating a dynamic progression system in AI-based educational chatbots through the embedding of Motivational Interviewing stages in our application.
We extended previous efforts in developing high-fidelity user simulations by introducing a novel implementation of \textit{cognitive model} system.
By using a different approach in generating cognitive model and introducing new descriptive fields, 
SWITCH simulates realistic social worker client interactions while allowing for key characteristics of each MI stage to affect the clients' behavior. 
Through the combination of skill score with LLM-based analytic methods, SWITCH introduces a dynamic progression management mechanism to control the MI stage.
Our study subsequently benchmarked the various methods of developing an accurate skill classification model to be used in our progression system, 
and our findings suggests that a BERT classifier model trained on transcripts of social worker interactions is a feasible method to accurately classify social worker messages with the skills used.

Future work can be done to create a more comprehensive training solution. 
Specifically, with the data of the social work skills used by the trainee, useful and personalized feedback can be generated either after each individual interaction, or when the training session is done.
Furthermore, the current solution also leaves room for future implementations of SWITCH in voice and video based systems.

\label{sec:conclusion}